\documentclass[10pt,twocolumn,letterpaper]{article}

\usepackage[pagenumbers]{cvpr}

\usepackage[T1]{fontenc}

\usepackage{amsmath}
\usepackage{amssymb}
\usepackage{booktabs}
\usepackage{graphicx}
\usepackage{xcolor}
\usepackage{tikz}
\usetikzlibrary{arrows.meta,positioning,calc}
\newcommand{\fitcol}[1]{\resizebox{\ifdim\width>\columnwidth\columnwidth\else\width\fi}{!}{#1}}
\newcommand{\fittext}[1]{\resizebox{\ifdim\width>\textwidth\textwidth\else\width\fi}{!}{#1}}

\IfFileExists{tables/numbers.tex}
  {
\newcommand{\BaselineSpec}{0.5011}
\newcommand{\DeleteCost}{-0.00080}
\newcommand{\SpecAloneMatched}{0.9360}
\newcommand{\SpecInFusionMatched}{0.5056}
\newcommand{\MatchedGap}{+0.4304}
\newcommand{\UnmatchedGap}{+0.4349}
\newcommand{\ConfoundWorth}{0.0045}

\newcommand{\SitePrior}{0.7871}

\newcommand{\FixMinusAvg}{+0.0020}
\newcommand{\FixMinusAvgP}{0.11}

\newcommand{\SpecParamPct}{72.1}
\newcommand{\CZeroBand}{0.90}
\newcommand{\CueSeeds}{three}
\newcommand{\CueSubStrength}{0.825}
\newcommand{\CueFlipStrength}{0.966}
\newcommand{\CueTopStrength}{0.995}
\newcommand{\CueNaturalTask}{0.943}
\newcommand{\ProbeDim}{192}
\newcommand{\ProbeC}{1.0}

\newcommand{\DeleteFloor}{0.01}
\newcommand{\SpecFeatureMean}{0.5437}
\newcommand{\FeatInputZeroSpec}{0.4522}

\newcommand{\FeatUntrainedMean}{0.4645}

\newcommand{\MatchedBand}{0.35}

\newcommand{\NBands}{28}
\newcommand{\NBandsHeld}{27}
\newcommand{\NFalsifiers}{5}

\newcommand{\SpecAloneBaseline}{0.9036}
\newcommand{\SpecAloneBaselineSD}{0.0623}
\newcommand{\SpecAloneMatchedSD}{0.0051}
\newcommand{\SpecAloneStuckSeeds}{2}
\newcommand{\SpecAloneStuckAUC}{0.790}

\newcommand{\SpecAloneMatchedLo}{0.929}
\newcommand{\SpecAloneMatchedHi}{0.942}

\newcommand{\WaveAloneMatched}{0.9400}
\newcommand{\ViewGapMatched}{+0.0040}
\newcommand{\CapBandLo}{0.80}

\newcommand{\UpdateRatioMean}{4.5}

\newcommand{\UpdateRatioEarly}{8.3}

\newcommand{\UpdateRatioLate}{4.9}
\newcommand{\UpdateRatioWindow}{19}

\newcommand{\SpecBestInFusion}{0.643}
\newcommand{\SpecBestInFusionMax}{0.696}

\newcommand{\NRecPN}{3{,}240}

\newcommand{\IcbhiCeiling}{0.6340}
\newcommand{\IcbhiBranchTol}{0.01}

\newcommand{\MelAxisMax}{0.008}
\newcommand{\MelAxisBand}{0.10}

\newcommand{\MelEmptyPct}{69.6}

\newcommand{\NOfficialVal}{301}

\newcommand{\SampleRateKHz}{22.05}
\newcommand{\WindowSec}{5}
\newcommand{\NFFT}{2048}
\newcommand{\HopLength}{512}
\newcommand{\NMels}{224}
\newcommand{\ImgSize}{224}
\newcommand{\TrainFracPN}{80/20}

\newcommand{\AudioFuseFused}{0.8608}

\newcommand{\AudioFuseViT}{0.8066}
\newcommand{\AudioFuseCNN}{0.8223}
\newcommand{\AudioFuseNParams}{2.56}

\newcommand{\NParamsM}{2.59}

\newcommand{\HeldoutDeleteCost}{-0.00005}
\newcommand{\HeldoutGap}{+0.4435}
\newcommand{\HeldoutGapSD}{0.0553}
\newcommand{\HeldoutSpecAlone}{0.9303}
\newcommand{\HeldoutSpecInFusion}{0.4868}

\newcommand{\Optimism}{+0.0039}

\newcommand{\OptimismArms}{10}
\newcommand{\InitAloneSpec}{0.8964}
\newcommand{\FullRecipeSpec}{0.9339}
\newcommand{\FrozenRandomSpec}{0.4061}
\newcommand{\InitAlonePct}{91}
\newcommand{\ProbeFloorV}{0.7903}
\newcommand{\ProbeInFusionV}{0.7995}
\newcommand{\ProbeDeltaV}{+0.0092}
\newcommand{\ProbeAloneV}{0.9164}
\newcommand{\ProbeInitOnly}{0.9084}
\newcommand{\ProbeInitOnlySD}{0.0083}
\newcommand{\ProbeDeltaCILo}{-0.0139}
\newcommand{\ProbeDeltaCIHi}{+0.0324}
\newcommand{\ProbeDeltaP}{0.39}
\newcommand{\ProbeDeltaNeg}{3}
\newcommand{\ProbeNSeeds}{10}
\newcommand{\ProbeAloneDeltaV}{+0.1261}
\newcommand{\ProbeAloneDeltaP}{0.001}
\newcommand{\ProbeAloneCILo}{+0.1123}

\newcommand{\ProbeBoundPct}{29}
\newcommand{\OgmGeSpec}{0.6076}

\newcommand{\LinearBaseSpec}{0.5475}

\newcommand{\RhoEpochOne}{82}
\newcommand{\OgmAlpha}{0.5}

\newcommand{\FusedMinusUnimodal}{+0.0003}
\newcommand{\FusedMinusUnimodalP}{0.87}

\newcommand{\NTrainV}{2{,}267}
\newcommand{\NValV}{486}
\newcommand{\NTestV}{487}

\newcommand{\CirCorBaseline}{0.7510}
\newcommand{\CirCorSpecAlone}{0.7988}
\newcommand{\CirCorSpecAloneTest}{0.7169}
\newcommand{\CirCorMargin}{+0.0477}
\newcommand{\CirCorBar}{+0.05}

\newcommand{\CirCorLinearView}{0.7269}

\newcommand{\CirCorGap}{+0.1531}

\newcommand{\CirCorGapLo}{+0.0133}
\newcommand{\CirCorGapHi}{+0.2807}
\newcommand{\CirCorSpecInFusion}{0.5638}

\newcommand{\CirCorProbeDelta}{+0.0120}

\newcommand{\CirCorProbeAloneDelta}{+0.0691}
\newcommand{\CirCorInitPct}{88.4}

\newcommand{\CirCorNHeld}{9}
\newcommand{\CirCorNMissed}{5}

\newcommand{\InitAlonePctTest}{88}

}
  {\typeout{^^J*** tables/numbers.tex not found: number macros are UNAVAILABLE. ***^^J}}

\definecolor{cvprblue}{rgb}{0.21,0.49,0.74}
\usepackage[pagebackref,breaklinks,colorlinks,allcolors=cvprblue]{hyperref}
\hypersetup{pdftitle={Strong Fused Performance Can Mask Branch Collapse in a Dual-View Heart-Sound Classifier: A Three-Part Diagnostic Protocol},pdfauthor={Safayet Khan, Hasan Zohirul Islam, Md. Tauhid Bin Iqbal}}

\title{Strong Fused Performance Can Mask Branch Collapse in a Dual-View Heart-Sound Classifier: A Three-Part Diagnostic Protocol}

\author{%
\begin{minipage}[t]{0.32\textwidth}\centering
Safayet Khan$^{*\dagger}$\\
{\tt\small ratulme12@gmail.com}
\end{minipage}%
\begin{minipage}[t]{0.32\textwidth}\centering
Hasan Zohirul Islam$^{*}$\\
{\tt\small hzihimel@gmail.com}
\end{minipage}%
\begin{minipage}[t]{0.32\textwidth}\centering
Md.\ Tauhid Bin Iqbal\\
{\tt\small tauhid.iqbal@ewubd.edu}
\end{minipage}%
\\ \noalign{\vspace{1.4em}}
Department of Computer Science and Engineering\\
East West University, Dhaka, Bangladesh
}

\begin{document}

\twocolumn[{%
\renewcommand\twocolumn[1][]{#1}%
\maketitle
\begin{center}
  \captionsetup{type=figure}
  \centering
  \includegraphics[width=\textwidth]{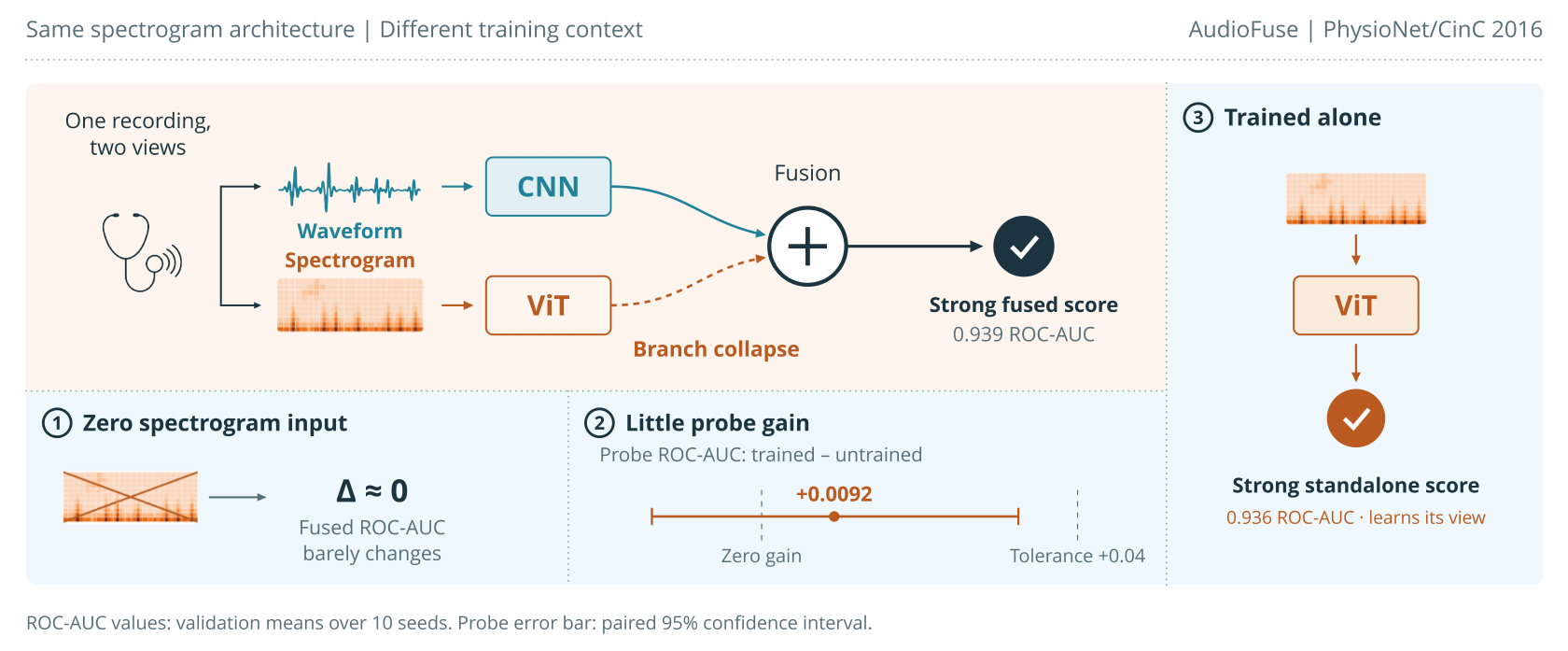}
  \caption{Strong fused performance can hide an unused branch (AudioFuse, PhysioNet/CinC 2016).
  (1)~Deleting the spectrogram branch barely changes fused ROC-AUC; (2)~a linear probe finds its
  features no better than an untrained copy's; (3)~trained alone, the same architecture learns the view.}
  \label{fig:teaser}
\end{center}
}]

{\renewcommand{\thefootnote}{\fnsymbol{footnote}}
\footnotetext[1]{Equal contribution.}
\footnotetext[2]{Corresponding author.}}

\begin{abstract}
Fusion models are usually judged by their fused score, a practice that assumes that every branch has learned something useful. We show that this assumption can fail silently in dual-view heart-sound classification. In AudioFuse, which fuses a spectrogram transformer with a waveform CNN, the spectrogram branch is unused despite strong fused performance when both branches are trained from scratch. We introduce a three-part diagnostic for branch collapse: whether deleting a branch changes the fused score, whether a linear probe on its features beats an untrained copy of the same branch, and whether the branch can learn its view when trained alone. Deleting the branch has negligible effect, and probing its features shows little improvement over an identically initialised, untrained copy. Yet the same architecture trained alone exceeds 0.9 AUC, so its view is learnable, while the jointly trained model is near chance when it must rely on this branch. Update statistics and an initialisation experiment point to an early optimisation imbalance. Moving a strong artificial label cue from one input to the other can switch which branch the model relies on, even though the branch architectures stay the same. We observe branch collapse on PhysioNet, but the full diagnostic is not met on ICBHI or CirCor. Fused performance alone therefore cannot establish that every branch learned a meaningful representation, and we recommend reporting branch deletion and an untrained-control probe alongside it.
\end{abstract}

\section{Introduction}
\label{sec:intro}

Fusion models combine multiple views of an input, but a strong fused score does not show whether every  branch  contributes. If one branch performs well alone, another can remain unused without an obvious  change in loss  curves or validation performance. We study this problem in  AudioFuse~\cite{siddiqui2026audiofuse}, a heart-sound classifier that combines a spectrogram transformer  with a waveform convolutional network.

Related work has studied such failures as modality imbalance and modality  competition~\cite{wang2020gradblend,wu2022greedy,peng2022ogm,huang2022competition}. Here, both views come  from the same recording, although preprocessing can preserve different information. An unused branch may  still contain useful features that the fusion head ignores. Contribution measures alone therefore cannot establish whether joint training improved its representation.

This distinction motivates three complementary checks (Section~\ref{sec:definition}; Supplementary Section~B.1 explains how the limits were set). We set a branch's input to zero to measure its effect on fused ROC-AUC (the \emph{deletion test}). We then compare linear probes~\cite{alain2017probes} on its frozen  features and on an identically initialised, untrained copy, testing whether joint training improved linear  decodability (the \emph{probe test}). Finally, we train the same architecture independently to establish  whether it can learn from its view (the \emph{standalone arm}). We diagnose \emph{branch collapse} only when  the deletion test shows negligible effect, the probe test shows no meaningful improvement over the untrained  copy, and the standalone arm reaches a specified performance level. The probe result does not establish  that nothing was learned.

Our reimplementation of AudioFuse meets all three criteria on PhysioNet/CinC 2016~\cite{liu2016physionet,clifford2017advances}.
Figure~\ref{fig:teaser}  summarises the three tests and the main findings. On the held-out test set, zeroing the spectrogram input has  negligible effect on fused ROC-AUC. Under matched training recipes, the standalone spectrogram architecture  reaches a test ROC-AUC of \HeldoutSpecAlone, whereas the jointly trained model reaches \HeldoutSpecInFusion\  when its waveform input is zeroed. The probe, which uses validation data only, stays near the untrained  copy.

We then test whether changes to training can restore branch use. A factorial experiment separates the  effects of starting weights, early freezing, and branch learning rate. Initialisation alone, using  independently trained branch weights, accounts for \InitAlonePctTest\% of the full recipe's gain in  spectrogram-only held-out ROC-AUC. Together with the waveform branch's weights changing several times more per epoch early in training, this is consistent with an optimisation imbalance. Branch use is restored without a demonstrated fused ROC-AUC benefit over score averaging.

Our contributions are:
\begin{itemize}
\item A case of collapse in a dual-view model, where both inputs come from the same recording: the larger
branch can be deleted at no cost and is no better than an untrained copy, yet learns its view when
trained alone.
\item A diagnostic for branch collapse that combines the deletion test, a probe read against an
identically initialised untrained copy, and the standalone arm; on two further corpora it withholds the
diagnosis, and on one it prevents a call that the first two tests alone would make.
\item Experiments on what decides branch use. Starting the branches from independently trained weights
restores the spectrogram branch, whereas the other changes we test restore it at most partly. When a
strong synthetic cue moves to the other view, the model switches to that branch, so the weaker branch
follows the signal, not the architecture.
\end{itemize}

Our study is narrow. We examine one architecture family trained from scratch, and we observe
collapse on one corpus, PhysioNet/CinC 2016. On ICBHI 2017~\cite{rocha2019icbhi} and CirCor
DigiScope~\cite{oliveira2022circor,reyna2023murmur}, the diagnostic does not call collapse.
Sub-database identity in PhysioNet predicts the label even without audio, so we read absolute
scores with caution and compare models on the same recordings. Within these limits, a strong
fused score is not enough on its own, and branch-level measurements should be reported
alongside it.

\section{Related Work}
\label{sec:related}

\noindent\textbf{Modality imbalance and competition.}
During joint training, the model may rely mainly on one modality while the others remain under-trained~\cite{wu2022greedy,peng2022ogm,du2023unimodal}. The fused model can also perform worse than the best model trained on a single modality~\cite{wang2020gradblend}. Huang et al.~\cite{huang2022competition} analyse a simplified model and show that competition can prevent task-relevant feature learning, with random initialisation influencing which modalities are learned. OGM-GE and related methods rebalance gradients~\cite{peng2022ogm,wei2024bml,li2023agm}. Other approaches use class prototypes~\cite{fan2023pmr}, unimodal objectives~\cite{wei2024mmpareto}, alternating adaptation~\cite{zhang2024mla}, or modality dropout~\cite{neverova2016moddrop}. Tang et al.~\cite{tang2025initialstate} use unimodal training to shape the initial states before joint training. BalanceBenchmark~\cite{xu2025balancebenchmark} compares methods under a common protocol.

\noindent\textbf{Dual-view heart-sound models.}
AudioFuse~\cite{siddiqui2026audiofuse} reports a fused ROC-AUC above either branch trained alone. Liu et
al.~\cite{liu2026timefreq} combine waveform and Mel-spectrogram features through cross-attention. Ranipa et
al.~\cite{ranipa2025aftvit} use two transformer streams with different patch sizes on the same cepstral map
to limit domination by the stream with more tokens. These designs and performance gains do not establish
that both branches contribute. CardioPHON~\cite{despotovic2026cardiophon} and the models of
Mathew et al.~\cite{mathew2024foundation} use self-supervised pretraining. Our models use no external
pretraining; to restore branch use, we initialise the branches with weights learned independently from
scratch on the same corpus. Whether collapse persists with self-supervised pretrained encoders remains
untested.

\noindent\textbf{Branch utilisation and probing.}
Shapley-based scores~\cite{xu2025balancebenchmark} and permutation importance~\cite{singh2024utilization}
estimate contribution to fused predictions. Low contribution may reflect limited feature learning (Huang et
al.~\cite{huang2022competition}) or useful features masked by fusion-head interactions (Chaudhuri et
al.~\cite{chaudhuri2025collapse}). Input ablation tests reliance, while linear
probes~\cite{alain2017probes} assess the branch's representation. Together, they can help distinguish these
explanations when task information is linearly accessible. Huang et al.\ already probe frozen jointly
trained encoders. Random-encoder baselines~\cite{zhang2018lmsyntax,wieting2019random} and control
tasks~\cite{hewitt2019control} provide precedents for interpreting probe scores against references. Among
the branch-utilisation studies reviewed here, we found no probe comparison with an initialisation-matched
untrained encoder. Our diagnostic combines this reference with the deletion test and the standalone arm.

\section{Setup}
\label{sec:setup}

\subsection{Data and preprocessing}
\label{sec:data}
\label{sec:materials}
\label{sec:data-physionet}
\label{sec:data-icbhi}
\label{sec:data-circor}
\label{sec:preprocessing}

PhysioNet/CinC 2016 is our main corpus; ICBHI 2017 and CirCor DigiScope test whether the diagnostic can withhold the diagnosis.
Table~\ref{tab:setup} summarises the data and splits.

PhysioNet's public training release~\cite{liu2016physionet,clifford2017advances} contains \NRecPN\
recordings labelled normal or abnormal across six sub-databases. Without patient identifiers, we guarantee
recording-level separation only. Our audit found all \NOfficialVal\ official validation recordings in
the training release (Supplementary Section~A.1). We create custom partitions to limit duplicate
leakage~\cite{kapoor2023leakage,barz2020duplicates}. These differ from AudioFuse's
reported patient-level split~\cite{siddiqui2026audiofuse}, preventing direct comparison of absolute scores.
We use an \TrainFracPN\ development split stratified by sub-database and label, and a content-hashed
partition grouping identical recordings before splitting. Unless stated
otherwise, PhysioNet results use the content-hashed validation set. Its test set is excluded from model
selection and scored once per planned checkpoint. Sub-database identity alone predicts the label at
\SitePrior\ AUC on the development split (Supplementary Section~F.1), so our central claims rest on
contrasts between arms scored on identical recordings.

ICBHI respiratory cycles~\cite{rocha2019icbhi} are labelled abnormal when crackles or wheezes are present.
For CirCor~\cite{oliveira2022circor,reyna2023murmur}, recordings from murmur-absent patients are negative;
those from annotated murmur locations in murmur-present patients are positive. We exclude recordings from
their other locations and from patients with unknown murmur status. We split both corpora by patient
to assess generalisation to unseen patients~\cite{saeb2017usecase}; detailed labelling rules are in
Supplementary Sections~A.2 and~A.3.

ICBHI recordings are first resampled to 4 kHz to equalise bandwidth across devices. CirCor's native rate is
4 kHz; PhysioNet's is 2 kHz. Inputs are resampled to \SampleRateKHz\,kHz; both branches use the first
\WindowSec\,s of each recording or ICBHI respiratory cycle. Shorter inputs are zero-padded. PhysioNet needs
no padding: its shortest recording lasts 5.3\,s.

We compute Mel power spectrograms with librosa~\cite{mcfee2015librosa} using a \NFFT-point FFT,
\HopLength-sample hop, and \NMels\ Mel bands. Each is converted to decibels relative to its maximum power,
resized to $\ImgSize \times \ImgSize$, and min--max normalised. This removes absolute signal level, which
the waveform retains (Section~\ref{sec:res-optimisation}). The filterbank is capped at 2\,kHz for ICBHI and
the main CirCor runs. PhysioNet retains AudioFuse's uncapped filterbank, with librosa's default upper limit
of 11.025\,kHz. \MelEmptyPct\% of its Mel filters lie entirely above PhysioNet's 1\,kHz band
limit. On CirCor, capping the filterbank changed the key comparisons by at most \MelAxisMax, within the pre-specified tolerance of $\pm\MelAxisBand$.

\begin{table*}[!t]
\centering
\caption{The three corpora as used. The first count row gives each corpus's development split; the second gives PhysioNet's content-hashed partition, whose validation portion is used in the main results tables. ICBHI and CirCor are split by patient; PhysioNet publishes no subject identifiers, so recording-level separation is the strictest standard available there.}
\label{tab:setup}
\small
\fittext{\begin{tabular}{lccc}
\toprule
 & PhysioNet/CinC 2016 & ICBHI 2017 & CirCor DigiScope \\
\midrule
Unit of classification & recording & respiratory cycle & recording \\
Development: train / val / test & 2592 / 648 / --- & 5360 / 1538 / --- & 1991 / 431 / 425 \\
Content-hashed: train / val / test & \NTrainV\ / \NValV\ / \NTestV & --- & --- \\
Positive fraction (train) & 0.205 & 0.484 & 0.170 \\
Positive fraction (val) & 0.207 & 0.430 & 0.176 \\
Split unit & recording & \textbf{patient} & \textbf{patient} \\
Subjects in validation & unknown & 26 & 131 \\
\bottomrule
\end{tabular}
}
\end{table*}

\subsection{Model}
\label{sec:model}
\label{sec:model-arch}

\begin{figure}[!t]
\centering
\includegraphics[width=\columnwidth]{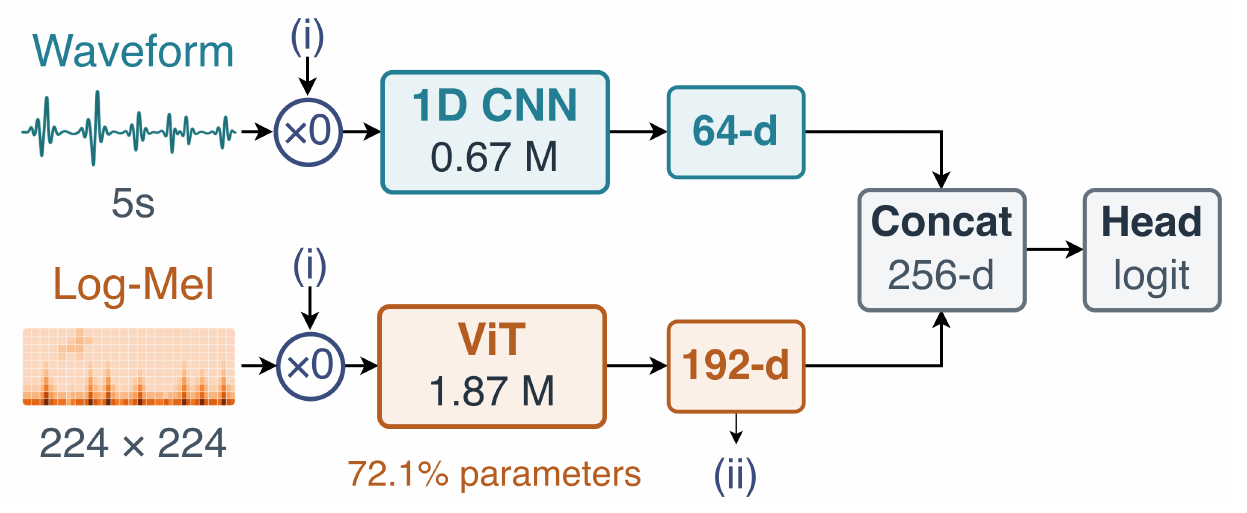}

\caption{AudioFuse as reimplemented, with the two diagnostic measurements. (i) Input ablation sets one
input to zero, marked $\times 0$, during evaluation. (ii) The linear probe compares frozen spectrogram
features with those from an identically initialised, untrained copy of the branch. Section~\ref{sec:protocol}
defines both procedures.}
\label{fig:arch}
\end{figure}

We reimplement AudioFuse's concatenation variant, which uses feature-level fusion~\cite{baltrusaitis2019survey}: the
branches process their inputs separately before their features are combined. The spectrogram branch is a
six-block vision transformer~\cite{dosovitskiy2021vit} whose token representations are averaged into a
\ProbeDim-dimensional feature vector. The waveform branch is a three-block one-dimensional CNN with batch
normalisation~\cite{ioffe2015batchnorm}, producing 64-dimensional features. Concatenating these features
gives a 256-dimensional vector, which passes through a two-layer head with dropout~\cite{srivastava2014dropout}
to produce one logit. Fig.~\ref{fig:arch} shows the architecture and where the deletion and probe tests are
applied.

The model has \NParamsM\ million parameters, close to AudioFuse's \AudioFuseNParams\ million;
\SpecParamPct\% belong to the spectrogram branch, so the branch examined for collapse is the larger one. We
train with AdamW~\cite{loshchilov2019adamw} and binary cross-entropy, weighting the positive class by the
ratio of negative to positive examples. Layer details are in Supplementary Section~A.4.

\subsection{Training regimes}
\label{sec:model-regimes}

We compare five training regimes. \emph{Joint training} starts both branches from random weights and
optimises the fused loss. \emph{Single-branch training} gives each architecture a separate head and trains
it on its own view. \emph{Gradient modulation} reduces the gradient of the branch estimated to contribute
more~\cite{peng2022ogm}. We test an adaptation for the non-linear head without generalization enhancement
(GE), and faithful OGM-GE with a linear head ($\alpha = \sigma = \OgmAlpha$; untuned). \emph{Modality
dropout} zeroes one randomly chosen view with probability 0.5 per sample during
training~\cite{neverova2016moddrop}. \emph{Pretrained-branch initialisation} loads independently trained
branch weights into the fused model~\cite{tang2025initialstate}. Its full recipe adds initial freezing and a
reduced branch learning rate. A $2 \times 2 \times 2$ factorial separates these three factors
(Section~\ref{sec:res-whatfix}).

The joint reference uses the \emph{baseline} recipe. On content-hashed PhysioNet it uses
\emph{baseline-AUC}, which changes only checkpoint selection, from validation accuracy to validation
ROC-AUC. The OGM adaptation and modality dropout keep accuracy selection, with joint training under the
baseline recipe as their reference. Faithful OGM-GE selects on AUC against its linear-head reference. The
standalone spectrogram model uses the \emph{spectrogram recipe}: a lower learning rate, linear
warm-up~\cite{goyal2017warmup}, class-balanced sampling, longer early-stopping patience, and checkpoint
selection and plateau scheduling based on validation AUC. Applying the spectrogram recipe to fusion tests
whether recipe differences explain the gap between the spectrogram branch trained alone and inside the
fused model.

Under baseline-AUC, two of ten standalone spectrogram runs stopped before beginning to fit the training
data. These failures contribute to the arm's variability without establishing a capacity limit.

A warm-up defect substantially limited fine-tuning in the original pretrained-initialisation trainer,
affecting the pretrained-init bars of Fig.~\ref{fig:ablation} and the corresponding held-out and
score-averaging results. The factorial uses no warm-up and a different reduced branch learning rate from
the original recipe (Supplementary Section~A.5); its initialisation-only checkpoint supplies the probe. Within
this factorial, initialisation alone accounts for \InitAlonePctTest\% of the combined setting's gain in
spectrogram-only held-out ROC-AUC.

We use ten seeds for content-hashed PhysioNet and CirCor, five for PhysioNet's development split and ICBHI,
and three per cue arm. Recipes, learning rates, checkpoint metrics, and stopping rules are in Supplementary
Section~A.5.

\subsection{Evaluation and analysis plan}
\label{sec:plan}

We report means and standard deviations across seeds unless stated otherwise. We compare arms on identical
recordings using paired $t$-tests on per-seed differences. Seed variability reflects optimisation on fixed
data; patient-clustered bootstrapping~\cite{efron1993bootstrap} estimates sampling variability where patient
identifiers are available. For pretrained initialisation versus probability averaging, we also use Wilcoxon
signed-rank and Yuen's trimmed $t$ checks. We treat $p$-values as descriptive because they are unadjusted
for multiple testing.

Plans for ICBHI, the matched protocol, held-out PhysioNet, and CirCor were committed before their runs.
They specify expected ranges (bands) and outcomes that would overturn a stated claim (falsifiers). For
example, the PhysioNet plan treats a held-out ROC-AUC gap below 0.15 between a branch trained alone and the
same architecture trained within fusion as evidence against collapse persisting on held-out data. Addenda
written during runs carry internal UTC timestamps in an append-only file and were committed with results;
the six probe bands were added before probing.
We report every band and falsifier, including those that missed, and identify unplanned analyses; complete
scored plans accompany the code. Supplementary Section~B describes the plans and statistical procedures.

\section{Diagnostic protocol}
\label{sec:protocol}
\label{sec:definition}

We diagnose branch collapse only when all three criteria below hold on data excluded from encoder training.
On PhysioNet, the deletion test and standalone arm use both validation and the held-out test set; the probe
uses validation only.

\begin{enumerate}
\item \textbf{Deletion test.} Zeroing the branch's input at inference changes fused ROC-AUC by less than
both the model's standard deviation across seeds and an absolute limit of \DeleteFloor.
\item \textbf{Probe test.} A linear probe on the branch's frozen features scores no more than 0.04 above the corresponding probe on an identically initialised, untrained copy. Both probes use the
same protocol. We call the untrained copy's probe score the probe floor.
\item \textbf{Standalone arm.} The same architecture, trained independently on its view, meets a
corpus-specific capability criterion. On PhysioNet this is a ROC-AUC of at least \CapBandLo. On CirCor
it is a margin of \CirCorBar\ over a conventional-feature baseline, fixed in advance; on ICBHI, where that
baseline was measured after the sweep, both branches fall within \IcbhiBranchTol\ of it
(Section~\ref{sec:scope}).
\end{enumerate}

The deletion test measures the branch's contribution to fused ROC-AUC under input zeroing. A small change
is consistent with limited reliance on that branch, even if it contains useful features. It does not
establish that the branch has no effect on individual predictions.

The probe test helps interpret this result. We treat a gain within tolerance as no meaningful improvement in
linear decodability under this protocol. This does not establish that the branch contains no information or
learned nothing. A larger gain, together with a negligible deletion effect, is consistent with useful
features that the fusion head underuses.

Standalone training shows that the view and architecture support learning under the tested recipe.
Applying that recipe to the fused model separately addresses the recipe mismatch
(Section~\ref{sec:model-regimes}). If the standalone criterion is not met, we do not diagnose collapse.
Deletion and probing use existing encoder checkpoints; the standalone arm requires separate encoder
training.

On PhysioNet, the capability criterion uses the lower edge of a range that was pre-specified for the
standalone arm under the fused model's recipe. We apply it to the arm trained under its own recipe, and we
do not treat the upper edge of the range as a capability ceiling.

We apply all criteria to means across seeds. Supplementary Section~B.1 gives the value of each limit and
tolerance and how it was chosen; Section~\ref{sec:plan} explains the plans' timing. The absolute deletion
limit was added after the experiments, ICBHI's standalone reference was measured afterwards, and the
PhysioNet capability criterion keeps only the lower edge of its pre-specified range. The probe tolerance was
set in an addendum informed by earlier development results. AudioFuse meets the deletion criterion without
the added absolute limit.

\section{Case study: AudioFuse on PhysioNet}
\label{sec:results}

We apply the three tests to AudioFuse on PhysioNet/CinC 2016. Unless stated otherwise, numbers are
validation ROC-AUC on the content-hashed partition (identical recordings kept together,
Section~\ref{sec:data}), averaged over ten seeds. The PhysioNet plan (Section~\ref{sec:plan}) fixed
\NBands\ bands and \NFalsifiers\ falsifiers in advance. Of these, \NBandsHeld\ bands held and no falsifiers fired. The one missed band is a check on the early update asymmetry
(Section~\ref{sec:res-optimisation}), and the six probe bands all held.
Each planned checkpoint was also scored once on the held-out test set (Supplementary Section~D.3), and
validation and test agree closely, with a mean difference of $\Optimism$ across $\OptimismArms$ trained configurations: three joint baselines, the two branches trained alone under
each recipe, and the three remedies. On the test set, the spectrogram architecture trained alone exceeds
the same branch inside the fused model by $\HeldoutGap \pm \HeldoutGapSD$, above the minimum of
\MatchedBand\ set in advance. The standalone arm reads $\HeldoutSpecAlone$, which meets the standalone
criterion, and deleting the branch changes the fused AUC by only $\HeldoutDeleteCost$.

\subsection{Deletion test: spectrogram branch is unused}
\label{sec:ablation}
\label{sec:res-unused}

\begin{figure}[!t]
\centering
\includegraphics[width=\columnwidth]{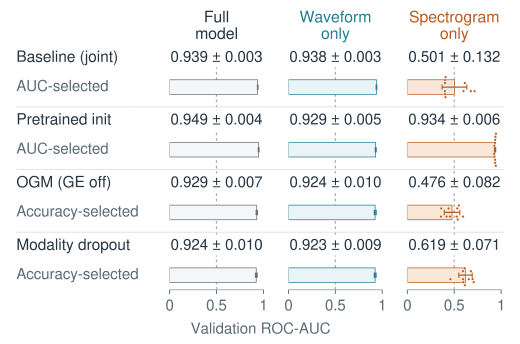}
\caption{Branch ablation on PhysioNet/CinC 2016. Bars show validation ROC-AUC as the mean over ten
seeds, and error bars show the standard deviation. Orange dots show each seed's spectrogram-only
score, spread out vertically so they do not overlap. For the waveform-only and spectrogram-only
scores, the other input is set to zero at test time. Dotted lines mark chance (0.5). The pretrained
init model uses the original warm-up (Section~\ref{sec:model-regimes}). Numerical values are in
Supplementary Section~D.1; the accuracy-selected joint baseline, which is the reference for OGM and
modality dropout, is in Section~D.3.}
\label{fig:ablation}
\end{figure}

The deletion test zeroes one branch's input in the trained model. The spectrogram branch holds
\SpecParamPct\% of the parameters, yet zeroing its input changes the full-model AUC by only $\DeleteCost$\
(Fig.~\ref{fig:ablation}, baseline regime). Zeroing the waveform input instead leaves a
spectrogram-only model near chance. The waveform branch alone matches the full model, and the fused model is no better than that branch trained by itself under the same recipe (held-out test set: 0.9412 against 0.9409, \FusedMinusUnimodal, $p=\FusedMinusUnimodalP$).

Zeroing is not a neutral intervention, because the model never saw an all-zero spectrogram in training.
We therefore also replaced the branch's pooled features with their training-set mean, and with a
permutation of them across the batch. For the spectrogram branch the score stays between
\FeatInputZeroSpec\ and \SpecFeatureMean, near chance under all three, and an untrained network scores near
\FeatUntrainedMean, so the procedure does not create the effect (five-seed development run; Supplementary
Section~C.1). The test shows that the model does not use the branch, not whether the branch learned
anything.

\subsection{Probe test: no gain over the untrained copy}
\label{sec:probe}
\label{sec:res-notlearned}

To measure linear decodability~\cite{alain2017probes}, we freeze the branch and extract its
\ProbeDim-dimensional pooled output for every validation recording. We fit an $L_2$-regularised logistic
regression in scikit-learn~\cite{pedregosa2011sklearn} ($C = \ProbeC$) under five-fold stratified
cross-validation. The probe reads only the stored
spectrogram, so no waveform signal can reach these features. The untrained copy shares the seed, which
pairs the comparison seed by seed; for a pretrained branch it is the same branch before fine-tuning.

\begin{figure}[!t]
\centering
\includegraphics[width=\columnwidth]{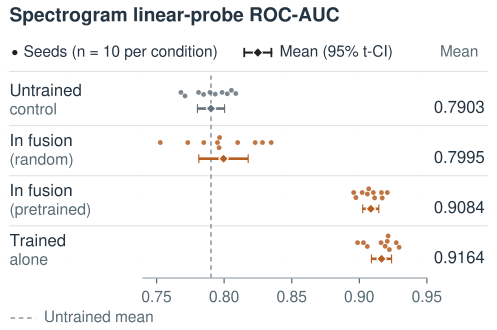}
\caption{Linear-probe ROC-AUC on the frozen features of the spectrogram branch (PhysioNet/CinC 2016;
five-fold cross-validation on the content-hashed validation set). Dots are the ten seeds of each
condition; diamonds and error bars are the means and their 95\% confidence intervals. The dashed line
is the untrained copy's mean, the probe floor. Inside the fused model the branch is no better than the
untrained copy; started from independently trained weights of the same seed, or trained alone, it is
well above it. The pretrained run uses no freezing and no warm-up (Section~\ref{sec:model-regimes}).}
\label{fig:probe}
\end{figure}

In fusion, the spectrogram branch reads \ProbeInFusionV, against the probe floor of \ProbeFloorV. The
difference is \ProbeDeltaV\ (95\% CI $[\ProbeDeltaCILo, \ProbeDeltaCIHi]$, paired $p = \ProbeDeltaP$),
which is indistinguishable from zero and within the tolerance of Section~\ref{sec:definition}
(Fig.~\ref{fig:probe}). The same architecture trained alone reads \ProbeAloneV, a gain of
\ProbeAloneDeltaV\ over the floor ($p < \ProbeAloneDeltaP$). The interval also bounds the effect: the
branch gains at most \ProbeDeltaCIHi\ in fusion, under \ProbeBoundPct\% of the gain of at least
\ProbeAloneCILo\ from training alone. The difference changes sign across seeds (\ProbeDeltaNeg\ of
\ProbeNSeeds\ fall below the floor), so we read it as no better than untrained, not as worse. The result
holds under three probe protocols (Supplementary Section~C.2).

With the deletion test, this is consistent with limited improvement in linear
decodability during joint training~\cite{huang2022competition}. It does not establish that nothing was learned.

\subsection{Standalone arm: the branch can learn the view}
\label{sec:standalone}
\label{sec:matched}
\label{sec:res-notcapacity}

The standalone arm trains the same architecture on the same recordings and split, with its own head and no fusion, over the same ten seeds. Under the recipe chosen for the branch, the spectrogram architecture reaches $\SpecAloneMatched \pm \SpecAloneMatchedSD$, which meets the standalone criterion (Section~\ref{sec:definition}), while the same branch inside the fused model scores \SpecInFusionMatched, near chance. The gap is not a capacity limit.

Under the fused model's recipe the standalone branch reaches $\SpecAloneBaseline \pm \SpecAloneBaselineSD$,
because \SpecAloneStuckSeeds\ of the ten runs stall at \SpecAloneStuckAUC. Under its own recipe every seed
falls between \SpecAloneMatchedLo\ and \SpecAloneMatchedHi, so we report the arm under that recipe. The
stalled runs were stopped early, not unable to learn: a linear probe on them still recovers the label
(unplanned check; Supplementary Section~C.3).

The two models were trained with different recipes, so the recipe might explain the gap. To test this, we
train the fused model with the branch's own recipe and delete the waveform input, the most favourable
setting for the spectrogram branch. The gap is \MatchedGap\ with the matched recipe and \UnmatchedGap\ with
the original one, so the recipe accounts for only \ConfoundWorth\ of it, and the gap is positive in every
seed. The matched model still has a waveform branch to rely on. The spectrogram branch is therefore unused, at the probe floor,
and capable of learning its view.

\subsection{Optimisation: an early update asymmetry}
\label{sec:res-optimisation}

One difference between the views is that spectrogram normalisation removes absolute level, which the
waveform keeps (Section~\ref{sec:preprocessing}). Even if this explained the whole standalone difference
between the views, \ViewGapMatched\ (\WaveAloneMatched\ against \SpecAloneMatched, both branches trained alone under the spectrogram recipe), that difference is two
orders of magnitude smaller than the collapse gap of \MatchedGap. This makes level an unlikely full
explanation at convergence. It does not address early training, which we did not test directly.

\begin{figure}[!t]
\centering
\includegraphics[width=\columnwidth]{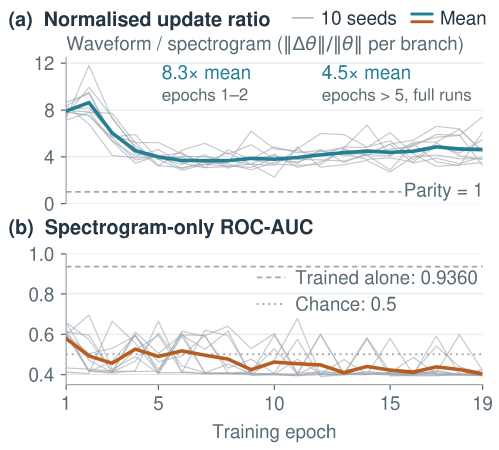}
\caption{Top: normalized parameter-update ratio (waveform branch $/$ spectrogram
branch), $\|\Delta\theta\|/\|\theta\|$ per epoch on PhysioNet/CinC 2016; the dashed line marks
parity. Bottom: the spectrogram branch's own AUC in those same runs, against its standalone
value and chance. Faint lines are the ten seeds, bold lines their means. The window is the
first \UpdateRatioWindow\ epochs, in which every seed is still training; runs that continue
sit slightly higher, at \UpdateRatioLate$\times$. The lower panel traces every epoch, whereas
Fig.~\ref{fig:ablation} reports the selected checkpoint.}
\label{fig:gradient}
\end{figure}

We track the per-epoch normalised parameter update $\|\Delta\theta\|/\|\theta\|$ within each branch, a per-epoch form of the learning speed of Wu et al.~\cite{wu2022greedy}. The waveform branch updates \UpdateRatioEarly$\times$ as much as the spectrogram branch over the first two
epochs, settling to \UpdateRatioMean$\times$ through convergence (Fig.~\ref{fig:gradient}). These are parameter updates, not gradients, and they do not show reliance, which the deletion test measures. By the OGM-GE estimator, the waveform branch leads on \RhoEpochOne\% of first-epoch steps, short of the \CZeroBand\ fraction set in advance. This is the one band that missed, so we place the asymmetry within the
first one or two epochs, not at the first step.

The spectrogram branch does not recover. In all ten runs, its spectrogram-only AUC peaks at
\SpecBestInFusion\ on average and never exceeds \SpecBestInFusionMax\ in any seed, against
\SpecAloneMatched\ when trained alone. The branch therefore does not learn the view and then lose
it, which agrees with the probe. Together with the asymmetry being largest in the first two
epochs, this suggests that collapse is settled early, before the spectrogram branch learns its
view. If so, the branch's starting weights should matter more than how it is trained afterwards.
The next subsection tests this.

\subsection{Independent initialisation restores the branch}
\label{sec:res-remedies}
\label{sec:res-whatfix}

Of the three remedies of Section~\ref{sec:model-regimes}, our gradient-modulation adaptation leaves the
spectrogram branch at baseline, and modality dropout lifts it only partway (Fig.~\ref{fig:ablation},
spectrogram-only bars). A faithful OGM-GE run lifts it only from \LinearBaseSpec\ to \OgmGeSpec\
(Supplementary Section~E.2), consistent with a related finding in medical image--text
learning~\cite{gu2026unimod}. Only pretrained-branch
initialisation restores the branch to essentially the standalone \SpecAloneMatched.

This remedy bundles three ingredients: independently trained starting weights, an early frozen phase and a
reduced branch learning rate. A $2\times2\times2$ factorial over the three (Table~\ref{tab:decompose})
shows that only the first carries the effect. Initialisation alone, with no freezing and the full learning
rate, restores the spectrogram branch to \InitAloneSpec, against \BaselineSpec\ at baseline and
\FullRecipeSpec\ for the full recipe, so it gives \InitAlonePct\% of the effect (\InitAlonePctTest\% on the
held-out test set). The other two ingredients add the small remainder, and freezing a randomly initialised
branch does not help (\FrozenRandomSpec).

\begin{table}[!t]
\centering
\caption{Decomposition of the pretrained-init recipe on PhysioNet/CinC 2016
(validation ROC-AUC, mean\,$\pm$\,s.d.\ over ten seeds). The $2\times2\times2$ varies
initialisation, the early frozen phase, and the branch learning rate.}
\label{tab:decompose}
\footnotesize
\setlength{\tabcolsep}{3pt}
\fitcol{\begin{tabular}{lllcc}
\toprule
Initialization & Frozen phase & Branch LR & Full AUC & \textbf{Spectrogram-only AUC} \\
\midrule
random & no & $1.0$ & 0.9388 $\pm$ 0.0029 & {0.5011 $\pm$ 0.1321} \\
random & no & $0.01$ & 0.9432 $\pm$ 0.0044 & {0.6697 $\pm$ 0.0402} \\
random & yes & $1.0$ & 0.9401 $\pm$ 0.0024 & {0.4061 $\pm$ 0.0152} \\
random & yes & $0.01$ & 0.9424 $\pm$ 0.0041 & {0.6868 $\pm$ 0.0491} \\
pretrained & no & $1.0$ & 0.9479 $\pm$ 0.0038 & {0.8964 $\pm$ 0.0635} \\
pretrained & no & $0.01$ & 0.9461 $\pm$ 0.0041 & {0.9313 $\pm$ 0.0082} \\
pretrained & yes & $1.0$ & 0.9495 $\pm$ 0.0041 & {0.9266 $\pm$ 0.0148} \\
pretrained & yes & $0.01$ & 0.9484 $\pm$ 0.0046 & {0.9339 $\pm$ 0.0070} \\
\bottomrule
\end{tabular}
}
\end{table}

A probe on the initialisation-only cell reads $\ProbeInitOnly \pm \ProbeInitOnlySD$, close to the standalone \ProbeAloneV\ and far above the floor of \ProbeFloorV. These features retain linearly accessible task information under a pretrained start. The small probe gain after joint training from random weights is consistent with limited feature learning~\cite{huang2022competition}. We read this
as evidence on the mechanism, not as a recommended training method.

Restoring the branch does not raise ROC-AUC beyond averaging the standalone branches. Rank-averaging
their predictions is at least as good as the restored fused model, and the fused model's edge over
probability-averaging is not significant ($\Delta = \FixMinusAvg$, paired $p = \FixMinusAvgP$; Supplementary
Section~E.3). The remedy restores use of the branch without a demonstrated gain in the task metric, so
performance and branch utilisation are separate properties.

\section{Mirrored-cue experiment}
\label{sec:cue}
\label{sec:scope-cue}

The measurements above show that a branch failed and could have succeeded, not what decides which branch
fails. Collapse may follow the view with the stronger learnable signal, or the architecture,
with the convolutional branch prevailing over the transformer. On PhysioNet the two cannot be separated,
because the waveform is both the convolutional branch's view and the view the model favours.

To separate them, we install a synthetic cue that carries the label in one view~\cite{kim2019learning,wu2022greedy}, retrain the model, and then mirror the cue into the other view and repeat. If the surviving branch follows the cue, collapse
follows the signal; if the same branch survives wherever the cue is placed, it follows the architecture.
Each cue is invisible to the other branch by construction, and the two cues carry equal information
(Supplementary Section~D.2). On PhysioNet we apply the cue at three strengths, given by the AUC of the cue
alone.

On ICBHI the result is symmetric. With the cue in the waveform, the spectrogram
branch is the weaker of the two in every seed, and with the cue in the spectrogram, the waveform branch is.
On PhysioNet the uncued fused model reaches \CueNaturalTask\ on the same validation recordings (five seeds), so an installed cue competes with the real task. A
spectrogram cue with an AUC of \CueSubStrength, below that, leaves the waveform branch dominant in every
seed, as in the uncued model, whereas cues with AUCs of \CueFlipStrength\ and \CueTopStrength, above it,
reverse the ordering in every seed.

For this pair of branches, the weaker branch is therefore not simply the transformer: it reverses when the
stronger signal moves between views, and the spectrogram view sustains the model once its signal is strong
enough. These arms use \CueSeeds\ seeds, and ICBHI cannot meet the standalone arm, so they show where
reliance moves, not collapse itself.

\section{When the diagnostic does not call collapse}
\label{sec:scope}

\subsection{CirCor: the standalone arm decides}
\label{sec:scope-circor}

On CirCor the view is informative: conventional descriptors reach \CirCorBaseline\ (waveform) and
\CirCorLinearView\ (spectrogram). The spectrogram branch does not clear the bar, however. Trained alone
under the matched recipe, it reaches \CirCorSpecAlone\ on validation, \CirCorMargin\ over the conventional
baseline and short of the \CirCorBar\ fixed in advance, and it also misses on the test set. The standalone
arm is therefore not met, so we do not call this case collapsed.

The asymmetry itself remains. Inside the fused model, the spectrogram branch reads \CirCorSpecInFusion,
against \CirCorSpecAloneTest\ for the same architecture trained alone, a gap of \CirCorGap\
[\CirCorGapLo, \CirCorGapHi] under a patient-clustered interval (Table~\ref{tab:circor}). The probe agrees
(\CirCorProbeDelta\ above its floor in fusion, \CirCorProbeAloneDelta\ alone), and independently trained
starting weights recover \CirCorInitPct\% of the distance. Of the pre-specified bands, \CirCorNHeld\ held
and \CirCorNMissed\ missed, and one of the five falsifiers fired (the standalone margin), which the plan
treats as grounds to withhold the diagnosis, not as a failed replication. Without the standalone arm, this case would
have been called collapsed.

\begin{table}[!t]
\centering
\caption{CirCor DigiScope on the held-out test set (ROC-AUC, mean\,$\pm$\,s.d.\ over ten seeds).
The spectrogram-only column is the fused model with the waveform input zeroed; the
single-branch arms have no such reading.}
\label{tab:circor}
\small
\fitcol{\begin{tabular}{lcc}
\toprule
Regime & Full model & Spectrogram only \\
\midrule
Waveform branch alone & 0.7724 $\pm$ 0.0118 & --- \\
Spectrogram branch alone & 0.7169 $\pm$ 0.0259 & --- \\
Fused model & 0.7782 $\pm$ 0.0161 & 0.5638 $\pm$ 0.0010 \\
Fused, pretrained-branch init & 0.7681 $\pm$ 0.0171 & 0.6997 $\pm$ 0.0294 \\
\bottomrule
\end{tabular}
}
\end{table}

\subsection{What the three corpora show}
\label{sec:scope-synthesis}

The two further corpora show where the diagnostic stops calling a branch collapsed. First, the view must carry
a learnable signal: on ICBHI both branches land within \IcbhiBranchTol\ of a conventional baseline at
\IcbhiCeiling, measured after the sweep, so no branch can meet the standalone arm. Second, the branch must
be able to learn that signal: on CirCor it does not clear its bar. Third, in our data the spectrogram
branch is used, and collapse does not occur, when the branches start from independently trained weights,
on PhysioNet (\InitAlonePctTest\%, Section~\ref{sec:res-whatfix}) and on CirCor (\CirCorInitPct\%), both
on held-out data. We saw collapse on one corpus, so we do not claim that these conditions are complete or
that they apply outside it.

One difference from AudioFuse remains open: it reports \AudioFuseFused\ fused against \AudioFuseViT\ and
\AudioFuseCNN\ for its branches alone, whereas our fused model is indistinguishable from its waveform branch on held-out data (\FusedMinusUnimodal, $p=\FusedMinusUnimodalP$). The split explains the difference in level but not
this one, and our runs cannot tell whether the cause is the split, the recipe, or collapse suppressing a
fusion benefit.

\section{Limitations}
\label{sec:threats}

We tested one architecture family trained from scratch, so whether the effect appears in other dual-view
models, or with self-supervised pretrained encoders, is untested. The mirrored-cue evidence rests on
\CueSeeds\ seeds, and the probe measures linear decodability only. The optimisation evidence is consistent
with an early update asymmetry and an initialisation effect, but we do not claim to have proved a cause.

\section{Conclusion}
\label{sec:conclusion}

A strong fused score does not show that both branches learned. In AudioFuse on PhysioNet/CinC 2016, the waveform branch is sufficient alone, so the fused score is strong while the spectrogram branch is unused, at the probe floor, and yet capable of learning its view. Nothing in a loss curve or a validation score reveals this. Starting the branches from independently trained weights restores the spectrogram branch without a demonstrated gain in fused ROC-AUC, and for this pair of branches the unused branch follows the weaker signal, not the architecture.

We recommend reporting two branch-level measurements alongside the fused score: the cost of deleting each branch, and a linear probe on its features read against an untrained copy. Both come from an existing checkpoint, and a branch that is free to delete and at the probe floor should then be trained alone. These results do not question the accuracy of published dual-branch models, but the claim that two views combine needs branch-level evidence.

\section*{LLM Usage Statement}
\label{sec:llm-usage}
We used Claude (Anthropic) to help refine the experimental plan, assist with code development, including plotting code, and edit the manuscript for grammar and clarity. The authors made the final experimental design decisions, reviewed and tested the LLM-assisted code, conducted all experiments, and verified the reported results. The authors verified all LLM-assisted content and take full responsibility for the paper.

\bibliographystyle{ieeenat_fullname}
\bibliography{refs}

\end{document}